\documentclass[10pt,conference]{IEEEtran}
\IEEEoverridecommandlockouts
\usepackage{cite}
\usepackage{amsmath,amssymb,amsfonts}
\usepackage{algorithmic}
\usepackage{graphicx}
\usepackage{textcomp}
\usepackage{xcolor}
\usepackage[acronym]{glossaries}
\usepackage{subfigure}
\usepackage{comment}
\newacronym{radar}{RADAR}{RAdio Detection And Ranging}
\newacronym{lidar}{LIDAR}{LIght Detection and Ranging}
\newacronym{lstm}{LSTM}{Long Short-Term Memory}
\newacronym{cnn}{CNN}{Convolutional Neural Network}
\newacronym{knn}{K-NN}{K-Nearest Neighbor}
\newacronym{mlp}{MLP}{Multi-Layer Perceptron}
\newacronym{tfn}{TFN}{Transformer Network}
\newacronym{mpnn}{MPNN}{Message Passing Neural Network}
\newacronym{svm}{SVM}{Support Vector Machine}
\newacronym{gflop}{GFLOP}{Giga Floating Point OPeration}
\newacronym{mb}{MB}{Mega Bytes}
\newacronym{ahc}{AHC}{Agglomerative Hierarchical Clustering}
\newacronym{iot}{IoT}{Internet of Things}
\newacronym{tfnet}{TFNet}{Transformer Network}
\newacronym{gpu}{GPU}{Graphical Processing Unit}
\newacronym{adc}{ADC}{Analog to Digital Conversion}
\newacronym{gnn}{GNN}{Graph Neural Network}
\newacronym{nlp}{NLP}{Natural Language Processing}
\newacronym{rnn}{RNN}{Recurrent Neural Network}
\newacronym{fmcw}{FMCW}{Frequency-Modulated Continuous Wave}
\newacronym{mimo}{MIMO}{Multiple-Input and Multiple-Output}
\newacronym{fft}{FFT}{Fast Fourier Transform}
\newacronym{cfar}{CFAR}{Constant False Alarm Rate}
\newacronym{ubpg}{UBPG}{Upper Body Point Cloud Gestures}
\newacronym{rf}{RF}{Radio Frequency}
\newacronym{dec}{DEC}{Dynamic Edge Convolution}
\newacronym{auc}{AUC}{Area Under ROC Curve}
\newacronym{rl}{RL}{Reinforcement Learning}
\newacronym{csi}{CSI}{Channel State Information}
\newacronym{ap}{AP}{Average Precision}
\newacronym{mmwave}{mmWave}{Millimeter Wave}
\newacronym{gnb}{gNB}{gNodeB}
\newacronym{ue}{UE}{User Equipment}
\newacronym{dqn}{DQN}{Deep Q-Networks}
\newacronym{trl}{TRL}{Transfer Reinforcement Learning}
\newacronym{a3c}{A3C}{Asynchronous Advantage Actor Critic}
\newacronym{gan}{GAN}{Generative Adversarial Network}
\newacronym{nr}{NR}{New Radio}
\newacronym{fr2}{FR2}{Frequency Range 2}
\newacronym{mac}{MAC}{Medium Access Control}
\newacronym{nzp}{NZP}{Non-Zero-Power}
\newacronym{csi-rs}{CSI-RS}{Channel State Information Reference Signal}
\newacronym{ssb}{SSB}{Synchronization Signal Blocks}
\newacronym{rsrp}{RSRP}{Reference Signal Received Power}
\newacronym{srs}{SRS}{Sounding Reference Signal}
\newacronym{gps}{GPS}{Global Positioning System}
\newacronym{macop}{MAC}{Multiply-ACCumulate}
\newacronym{ai}{AI}{Artificial Intelligence}
\newacronym{ran}{RAN}{Radio Access Network}
\def\BibTeX{{\rm B\kern-.05em{\sc i\kern-.025em b}\kern-.08em
    T\kern-.1667em\lower.7ex\hbox{E}\kern-.125emX}}
\begin{document}

\title{Environment-Aware Transfer Reinforcement Learning for Sustainable Beam Selection}

\author{Dariush~Salami,~Ramin~Hashemi,~Parham~Kazemi,~and~Mikko~A.~Uusitalo
\IEEEcompsocitemizethanks{

\IEEEcompsocthanksitem D. Salami and M. A. Uusitalo are with Nokia Bell Labs, Espoo,
Finland.\protect\\
E-mails: \{dariush.salami, mikko.uusitalo\}@nokia-bell-labs.com

\IEEEcompsocthanksitem R. Hashemi and P. Kazemi are with Nokia Standards, Espoo, Finland.\protect\\
E-mails: \{ramin.hashemi, parham.kazemi\}@nokia.com
}
}

\maketitle

\begin{abstract}
This paper presents a novel and sustainable approach for improving beam selection in 5G and beyond networks using transfer learning and \gls{rl}. Traditional \gls{rl}-based beam selection models require extensive training time and computational resources, particularly when deployed in diverse environments with varying propagation characteristics posing a major challenge for scalability and energy efficiency. To address this, we propose modeling the environment as a point cloud, where each point represents the locations of \glspl{gnb} and surrounding scatterers. By computing the Chamfer distance between point clouds, structurally similar environments can be efficiently identified, enabling the reuse of pre-trained models through transfer learning. This methodology leads to a 16$\times$ reduction in training time and computational overhead, directly contributing to energy efficiency. By minimizing the need for retraining in each new deployment, our approach significantly lowers power consumption and supports the development of \emph{green} and \emph{sustainable} \gls{ai} in wireless systems. Furthermore, it accelerates time-to-deployment, reduces carbon emissions associated with training, and enhances the viability of deploying \gls{ai}-driven communication systems at the edge. Simulation results confirm that our approach maintains high performance while drastically cutting energy costs, demonstrating the potential of transfer learning to enable scalable, adaptive, and environmentally conscious \gls{rl}-based beam selection strategies in dynamic and diverse propagation environments.

\end{abstract}
\glsresetall


\section{Introduction}
\begin{figure}
    \centering
    \includegraphics[width=0.8\linewidth]{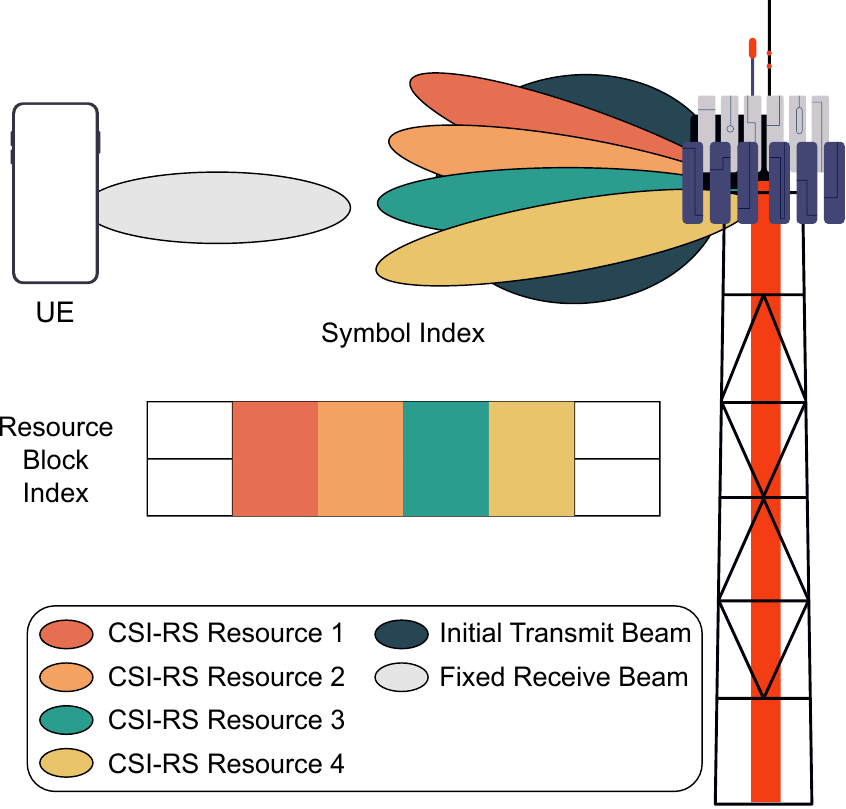}
    \caption{The transmit-end beam refinement procedure, considering four \gls{nzp}-\gls{csi-rs} resources transmitted in four different directions.}
    \label{fig:beam_selection}
\end{figure}

Beam selection in 5G and beyond represents a pivotal advancement in wireless communication technology, addressing the challenges posed by the use of higher frequency bands, such as \gls{mmwave}~\cite{gao2016near}. Unlike lower frequency bands used in previous generations, \gls{mmwave} frequencies offer significantly higher data rates but suffer from increased path loss and greater susceptibility to obstacles, necessitating advanced techniques for maintaining robust and efficient communication links. Beam selection involves choosing the optimal directional beam for communication between \gls{gnb} and \gls{ue}, leveraging highly directional antennas to focus the signal and enhance its strength. By efficiently selecting the best beam, 5G and beyond systems can mitigate the effects of path loss and blockages, thereby maintaining a reliable communication link even in challenging conditions. Fig.~\ref{fig:beam_selection} depicts the beam selection procedure~\cite{3gpp2017study} explained in detail in Section~\ref{section:proposed}.

\gls{rl} is essential for beam selection in 5G and beyond due to its ability to handle the dynamic and complex nature of modern wireless environments~\cite{hu2021joint}. Unlike traditional static methods, \gls{rl} continuously learns from real-time interactions, optimizing beam selection decisions based on changing conditions such as user mobility, interference, and obstacles. This leads to enhanced network efficiency, reliability, and user experience, making \gls{rl} a critical component for realizing the full potential of 5G and future wireless networks.

When a \gls{gnb} is deployed in a new environment with different physical characteristics, there is a need to re-train the \gls{rl} agent to ensure optimal beam selection~\cite{elsayed2020transfer}. The physical environment, including the presence of buildings and other obstacles, significantly impacts signal propagation. An \gls{rl} agent trained in one environment may not perform well in another due to these varying factors. Re-training allows the \gls{rl} agent to learn the specific characteristics of the new environment. However, this re-training process can be computationally intensive, leading to increased power consumption.

A transfer learning approach can significantly reduce the need for re-training when deploying a \gls{gnb} in a new environment by leveraging knowledge from a pre-trained \gls{rl} agent~\cite{wang2022deep}. Instead of starting the training process, the pre-trained model is fine-tuned with a smaller set of data relevant to the new environment. This reduces the data and computational resources required, which in turn lowers power consumption.

In this work, a mechanism based on Chamfer distance is proposed to identify similar environments and share \gls{rl} models between these environments. The Chamfer distance allows us to quantitatively compare different physical environments by analyzing key features such as obstacle distribution. By identifying environments with high similarity, pre-trained \gls{rl} models can be transferred from one environment to another, ensuring that the \gls{rl} agent can leverage existing knowledge to quickly adapt to the new setting. This approach not only accelerates the deployment process but also conserves computational resources and reduces power consumption associated with extensive re-training.

\section{Related Work}
In the context of 5G and beyond networks, beam selection is a critical aspect that significantly influences the efficiency and reliability of wireless communication. Recent research has increasingly focused on leveraging \gls{rl} techniques to optimize beam selection processes. Various studies have explored different \gls{rl} algorithms, such as Q-learning, \glspl{dqn}~\cite{ju2023deep, kim2023joint}, and policy gradient methods, to enhance beamforming decisions, reduce latency, and improve overall network performance. However, despite the effectiveness of these methods, there remains a significant challenge in generalizability when deploying \glspl{gnb} in new environments with different propagation characteristics, necessitating further research to ensure robust and adaptive performance across diverse environments.

To address the issue of generalizability in \gls{rl} applications, researchers have begun exploring the concept of \gls{trl}~\cite{zhu2023transfer}. \gls{trl} extends traditional \gls{rl} by enabling a model trained in one environment to adapt and perform effectively in different, previously unseen environments. This approach leverages knowledge transfer, where the learning from a source domain (original environment) is transferred to a target domain (new environment), thereby reducing the amount of training data and time required in the target domain. For example, a transfer learning approach with a pre-trained ResNet50 model has been proposed in~\cite{varma2022effective} to increase efficiency and reduce computational complexity. This approach, integrates ResNet50 with \gls{a3c}, significantly reducing training time.

However, common transfer learning techniques like fine-tuning are often inadequate, sometimes making it quicker to retrain models from scratch. Another research~\cite{gamrian2019transfer} addresses this limitation by decoupling the visual transfer task from the control policy, resulting in improved sample efficiency and transfer performance. Using unaligned \glspl{gan} for the visual mapping from the target to the source domain, the approach allows the control policy to be refined through imitation learning from imperfect demonstrations. This method is validated on synthetic visual variants of the Breakout game and on transferring skills between levels of the Nintendo game Road Fighter, showcasing its effectiveness in enhancing transferability in \gls{rl} tasks.

While the introduced approaches have shown success in visual tasks, this work aims to apply transfer learning to beam selection in 5G and beyond networks. The propagation environment, including surrounding buildings and structures, is modeled as a point cloud. Then, using the Chamfer distance, similar environments are identified where exchanging pre-trained models can significantly reduce training time and increase efficiency. This approach leverages the spatial characteristics of the environment to optimize beam selection, demonstrating the versatility and potential of transfer learning beyond traditional visual tasks.

\section{Proposed Scheme}
\label{section:proposed}
\begin{figure*}
    \centering
    \includegraphics[width=0.95\linewidth]{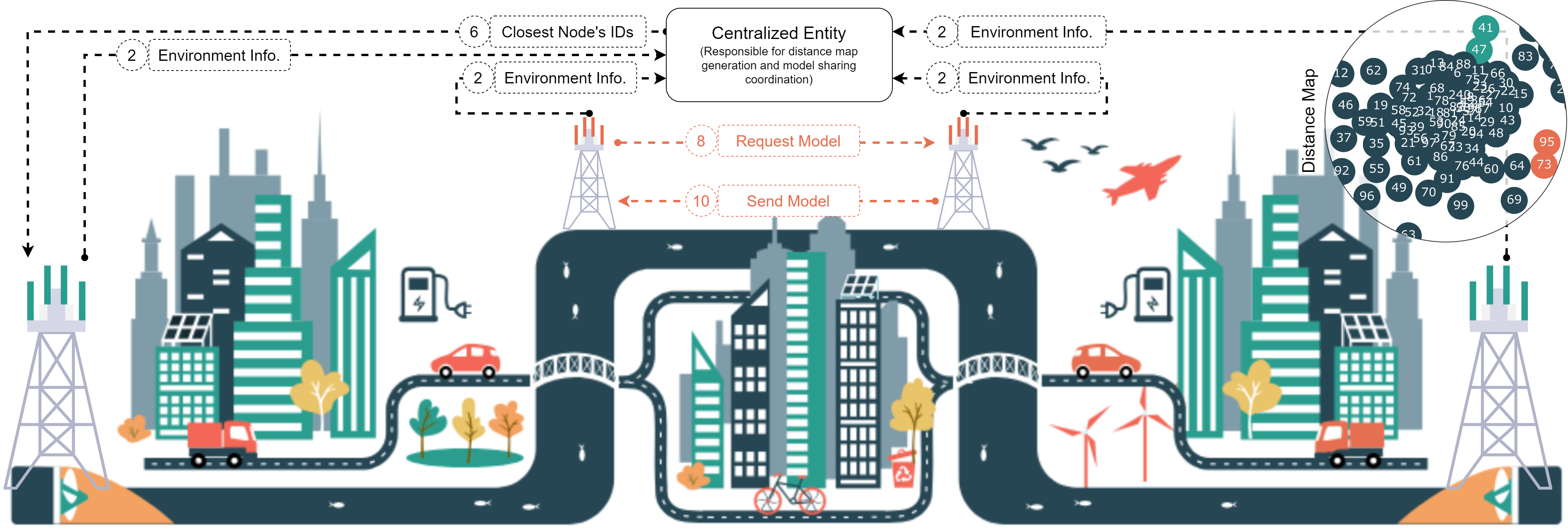}
    \caption{The schematic view of the proposed method. The \glspl{gnb} shown in the same colors (orange and green) have similar signal propagation characteristics (e.g. surrounding buildings). The numbers inside dashed circles refer to the steps in Fig.~\ref{fig:model_sharing}. Plus, a map of environments based on the pair-wise Chamfer distance is shown in top right circle. Orange nodes are the two environments with two orange \glspl{gnb}, and green ones are the ones with two green \glspl{gnb}.}
    \label{fig:teaser}
\end{figure*}

In \gls{nr} 5G, \gls{fr2} operates at \gls{mmwave} frequencies between 24.25 GHz to 52.6 GHz. At these higher frequencies, transmitted signals experience significant path loss and penetration loss, impacting the link budget. Beamforming is essential to enhance the gain and directionality of signal transmission and reception at these frequencies. Beam management, comprising Layer 1 (physical layer) and Layer 2 (\gls{mac}) procedures, is crucial for establishing and maintaining an optimal beam pair (transmit beam and corresponding receive beam) to ensure robust connectivity.

TR 38.802 Section 6.1.6.1~\cite{3gpp2017study} defines beam management as three procedures. Procedure 1 (P-1) involves initial acquisition using \gls{ssb} with beam sweeping at both transmit and receive ends to select the best beam pair based on \gls{rsrp} measurements. This initial wide transmit beam is shown in dark green in Fig.~\ref{fig:beam_selection}. Procedure 2 (P-2) focuses on transmit-end beam refinement, utilizing \gls{nzp}-\gls{csi-rs} for downlink and \gls{srs} for uplink, while keeping the receive beam fixed. This step refines the beams to achieve higher directivity and gain. The green beam shown in Fig.~\ref{fig:beam_selection} represents the best beam selected in this stage. Procedure 3 (P-3) addresses receive-end beam adjustment by sweeping the receive beam while maintaining the current transmit beam, using \gls{nzp}-\gls{csi-rs} for downlink and \gls{srs} for uplink. The best beams are selected based on \gls{rsrp} measurements.

However, the initial access beam selection involves exhaustive beam sweeping on both transmitter and receiver sides to find the beam pair with the strongest \gls{rsrp}. Given the numerous antenna elements and beams in \gls{fr2}, this exhaustive search is computationally expensive and increases initial access time. To mitigate this problem, an \gls{rl} agent can be used for beam selection by utilizing the \gls{gps} coordinates of the receiver and the current beam angle as the \gls{ue} moves. This approach reduces the need for repeated exhaustive searches and minimizes communication overhead.

However, the problem of generalizability in beam selection arises when a model trained in one environment is deployed in a new environment with different propagation patterns, such as variations in surrounding buildings. The \gls{rl} model may not perform well in the new setting due to these differences, leading to suboptimal beam selection and reduced signal quality. By modeling the environment as a point cloud and using this representation to find similar environments this problem can be mitigated. This approach facilitates transfer learning, allowing pre-trained models to adapt more effectively to new environments, thereby improving performance and reducing retraining time.

\subsection{Point Cloud Representation of Environments}
As shown in Fig.~\ref{fig:teaser}, different \glspl{gnb} might be deployed in similar environments where the signal propagation patterns are alike. In such cases, the control policies learned by \gls{rl} algorithms would also be similar. For instance, the red and green \glspl{gnb} pairs in Fig.~\ref{fig:teaser} are situated in comparable surroundings. To leverage this information, it is necessary to model the environment quantitatively to facilitate the comparison of different environments. The point cloud (e.g. a set of points in the space) representation has been widely used in the literature for different use cases~\cite{salami2022tesla,palipana2021pantomime}. In this work, the environment is modeled as a point cloud, with each point representing the location of \gls{gnb} and surrounding scatterers. The locations of these scatterers can be determined using \gls{gps}, directly using sensing capabilities in the network~\cite{xie2023environment, barneto2022millimeter}, or even third-party radar systems~\cite{salami2024angle}. Note that in this work, we do not consider the geometrical properties of the scatterers, such as their dimensions or material composition. However, it is certainly possible to extend the point clouds to a richer and higher-dimensional representation to incorporate these additional aspects.

To compare the similarity of different environments, the Chamfer distance is used to measure how closely two point clouds, each representing an environment, resemble each other. This method evaluates similarity by determining the shortest distance from each point in one environment to the closest point in the other, and vice versa. The overall similarity is assessed by averaging these minimal distances, providing a balanced comparison of the spatial characteristics of both environments.  

To compare the similarity of different environments, the Chamfer distance is proposed between two point clouds $P=\{p_1, p_2, ..., p_3\}$ and $Q=\{q_1, q_2, ..., q_3\}$ , representing two different environments:

\begin{equation}
    \begin{split}
        D_{CD}(P,Q) = \frac{1}{|P|}\sum_{p \in P}\operatorname{min}_{q \in Q}||p-q||^2 + \\ \frac{1}{|Q|}\sum_{q \in Q}\operatorname{min}_{p \in P}||q-p||^2,
    \end{split}
\end{equation}

\noindent where $D_{CD}$ is the Chamfer distance between two point clouds, indicating the similarity between two environments. It works by calculating, for each point in the first point cloud, the distance to the nearest point in the second cloud, and vice versa. These distances are then summed and averaged, capturing the extent to which the two sets of points cover each other.

The Chamfer distance has several key properties making it ideal for environment comparison. Symmetry ensures that the distance measurement is identical regardless of the order in which the two environments are compared. Non-negativity guarantees that the distance is always zero (identical environments) or positive. Additionally, density sensitivity allows it to capture fine-grained spatial details by considering the distribution of points, while its computational efficiency makes it scalable for large datasets making.

\subsection{Pair-Wise Distance Between Environments}
To generate a pair-wise distance map of the point cloud representations of different environments, we use the Kamada-Kawai method, a force-directed layout algorithm that arranges nodes (in this case, environments) based on pair-wise distances. Unlike the traditional approach where distances are based on graph topology, here we define the distance between environments using the Chamfer distance between their corresponding point clouds. The energy function minimized by Kamada-Kawai remains:

\begin{equation}
    E = \frac{1}{2} \sum_{i \ne j} k_{ij} \left( \|x_i - x_j\| - l_{ij} \right)^2
\end{equation}

where \( \|x_i - x_j\| \) is the layout distance between environments \( i \) and \( j \), \( l_{ij} \) is the target length set proportional to the Chamfer distance between their point clouds, and \( k_{ij} \) is typically set inversely proportional to that distance (e.g., \( k_{ij} = 1 / d_{ij}^2 \)). In other words, environments with similar spatial structures (low Chamfer distance) are placed closer together, while dissimilar ones (high Chamfer distance) are positioned further apart. This produces an interpretable spatial map that visually clusters similar environments.

The top-right circle in Fig.~\ref{fig:teaser} illustrates a sample pair-wise distance map. In this map, the green nodes correspond to two similar \glspl{gnb} on the left and right sides of the figure, while the red nodes represent two other similar pairs located in the middle. The map is generated from simulation data and visualized using the Kamada-Kawai method. As shown, the two pairs share similar environments, resulting in comparable signal propagation patterns and, consequently, smaller distances in the pair-wise distance map. In contrast, pairs with different colors are positioned farther apart on the map, indicating a lower degree of similarity between them.

\subsection{Model Sharing Scheme}
\begin{figure}
    \centering
    \includegraphics[width=\linewidth]{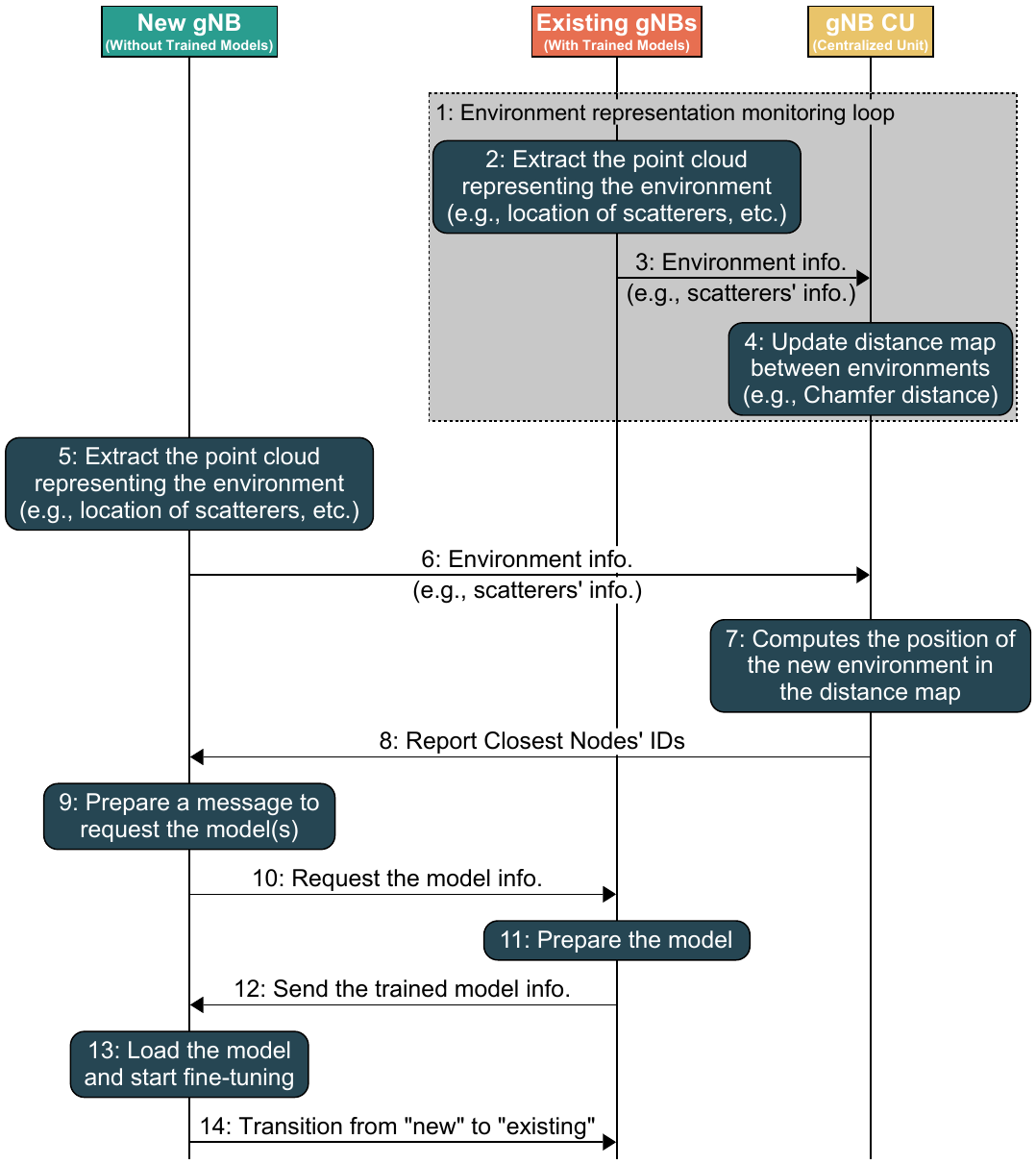}
    \caption{The sequence diagram of the proposed scheme for calculating the distance map and sharing the trained beam selection model accordingly.}
    \label{fig:model_sharing}
\end{figure}

Having generated the distance map, a mechanism is needed to share trained beam selection models between \glspl{gnb} deployed in similar environments. Fig.~\ref{fig:model_sharing} illustrates the sequence diagram for this model-sharing process. In the figure, the \textbf{New \gls{gnb}} is a recently deployed node without a trained model for beam selection. However, there are \textbf{Existing \glspl{gnb}} that have been deployed for some time and already possess trained beam selection models. Additionally, a \textbf{Centralized Unit}, which is one of the \textbf{Existing \glspl{gnb}}, is responsible for generating the distance map as detailed in Steps 4 and 7 of the diagram. As it is shown in the diagram, the existing \glspl{gnb} periodically extract the point cloud representations of their environments sharing them with the centralized unit to calculate the distance map. The centralized unit keeps and updated distance map using the information received from the existing \glspl{gnb}. 

A newly deployed \gls{gnb} extracts scatterer information and generates a point cloud representation. This information is shared with a centralized unit to identify nearby \glspl{gnb} with similar propagation characteristics using a pair-wise Chamfer distance map. The new \gls{gnb} then requests trained beam selection models from these \glspl{gnb}, receiving their model weights. While this work focuses on weight sharing, more sophisticated knowledge sharing approaches could be explored. After fine-tuning, the new \gls{gnb} contributes its state to the other \glspl{gnb} as shown in Step 14 of the signaling diagram.

\section{Implementation}
\label{sec:implementation}

For the beam selection environment, observations are represented by the \gls{ue} position information and the current beam selection. The actions involve selecting one of total beam angles from the \gls{gnb}. The reward $r_t$ at time step $t$ is:

\begin{equation}
    r_t = \underbrace{0.9\times\operatorname{rsrp}}_\text{\gls{rsrp} reward} + (\underbrace{-0.1\times|\theta_t - \theta_{t-1}|}_\text{penalty for control effort}),
    \label{eq:reward}
\end{equation}

$r_t$ is the reward based on the signal strength measured from the \gls{ue} (\gls{rsrp}) and includes a penalty for control effort, where $\theta$ is the beam angle in degrees. A high \gls{rsrp} value leads to a higher reward, promoting actions that improve signal quality. Conversely, frequent or large changes in beam angle are penalized, encouraging more stable control. This balance between signal strength and control effort helps to optimize the overall performance of the beam selection process.

\begin{figure}
    \centering
    \includegraphics[width=0.9\linewidth]{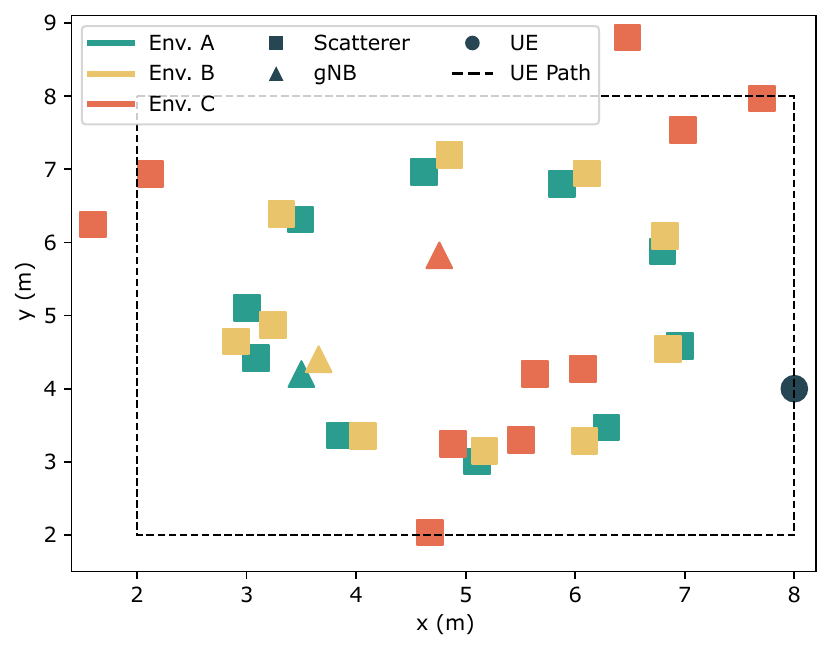}
    \caption{The three simulation environments are depicted in different colors. Environments A and B are more similar to each other because the locations of the scatterers and \glspl{gnb} are closer compared to environment C. In all environments, the \gls{ue} follows the path shown by the dashed square.}
    \label{fig:environment}
\end{figure}

To generate the simulation data, in the environment A shown in Fig.~\ref{fig:environment}, receivers are randomly distributed on the perimeter of a 6 m square and configured with 16 beam pairs (4 beams on each end, analog beamformed with one \gls{rf} chain). Using a \gls{mimo} scattering channel, 200 receiver locations in the training set and 100 receiver locations in the test sets in each environment are considered. The prerecorded data uses 2-D location coordinates. For environments B and C, random noise up to 0.25 m and 2 m is added to the locations of the \glspl{gnb} and scatterers, respectively. Moreover, given that receiver beam selection minimally impacts signal strength, the mean \gls{rsrp} is computed for each base station antenna beam at each \gls{ue} location. Therefore, the action space consists of 4 beam angles as shown in Fig.~\ref{fig:beam_selection}.

In the simulation, a \gls{dqn} agent is utilized to optimize beam selection decisions. The agent approximates the long-term reward based on observations and actions. This critic function represents a vector Q-value function where observations, such as \gls{ue} position and current beam selection, are inputs, and state-action values (Q-values) are outputs. Each output element of the critic corresponds to the expected cumulative long-term reward for selecting a specific beam angle from the current observation state. This allows the \gls{dqn} agent to iteratively learn and improve its beam selection strategy through reinforcement learning, aiming to maximize signal strength while minimizing control effort penalties.

\section{Evaluation}
This section evaluates the performance of the proposed scheme across different settings, examining its effectiveness in various environments. Detailed analyses are conducted to highlight the improvements in training time, computational efficiency, and overall system performance, demonstrating the robustness and versatility of the proposed approach.

\subsection{Model Performance vs Chamfer Distance}

\begin{figure}
    \centering
    \includegraphics[width=\linewidth]{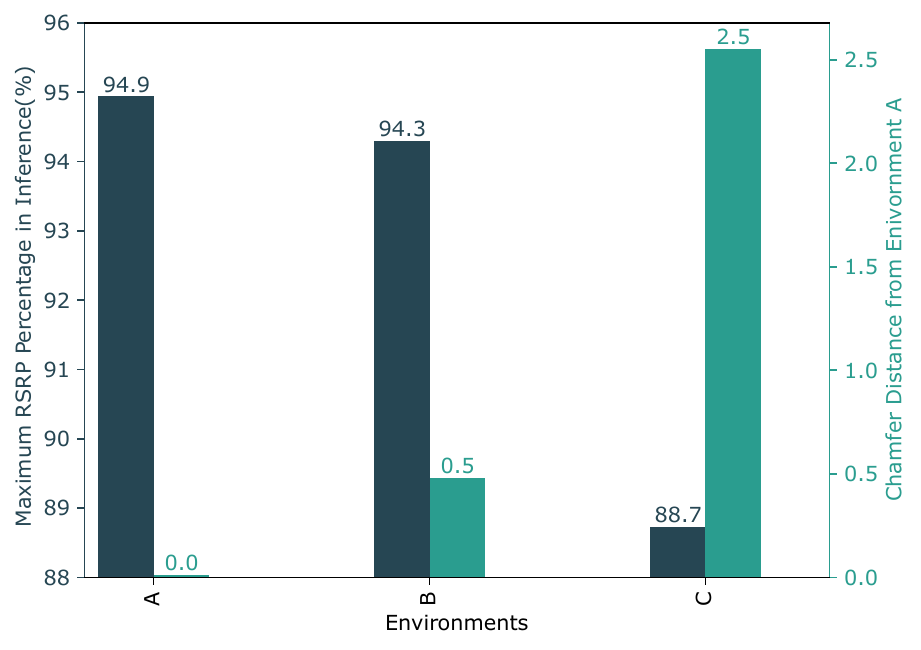}
    \caption{The left axis represents the ratio of the agent's \gls{rsrp} to the maximum \gls{rsrp}, while the right axis shows the Chamfer distance of the environments from environment A. All agents are trained on environment A and tested against environments A, B, and C as indicated on the x-axis.}
    \label{fig:performance_vs_chamfer}
\end{figure}

The evaluation metric used in this work is the ratio of the agent's \gls{rsrp} to the maximum \gls{rsrp}. In Fig.~\ref{fig:performance_vs_chamfer}, the left axis represents performance, while the right axis shows the Chamfer distance between the training environment (A) and the test environments (A, B, and C).

To evaluate the performance of the scheme under various conditions, the model is trained in environment A (See Section~\ref{sec:implementation} and shown in Fig.\ref{fig:environment}) and tested in all three environments. As illustrated in Fig.\ref{fig:performance_vs_chamfer}, the best performance is achieved when the model is tested in the same environment it was trained in. In this case, the agent's \gls{rsrp} reaches 94.9 $\%$ of the maximum \gls{rsrp}. However, when the model is used to perform beam selection in other environments, an inverse correlation is observed between the Chamfer distance of the training and test environments and the performance. For instance, in environment B, which has a Chamfer distance of 0.5 from environment A, the agent's \gls{rsrp} is slightly lower at 94.3 $\%$. In environment C, with a Chamfer distance of 2.5 from environment A, the agent's \gls{rsrp} drops to 88.7 $\%$. 

In other words, the more similar the environments, the higher the performance of the scheme. This demonstrates that utilizing transfer learning in similar environments is highly beneficial for beam selection using \gls{rl}. The significant drop in performance in environment C highlights the importance of considering the similarity of environments when applying transfer learning techniques.

\subsection{Performance Gain}

In this section, the results are presented to demonstrate the advantages of employing transfer learning for beam selection using \gls{rl}. It is shown that transfer learning not only enhances performance in similar environments but also substantially reduces training time and computational requirements.

\begin{figure}
    \centering
    \includegraphics[width=\linewidth]{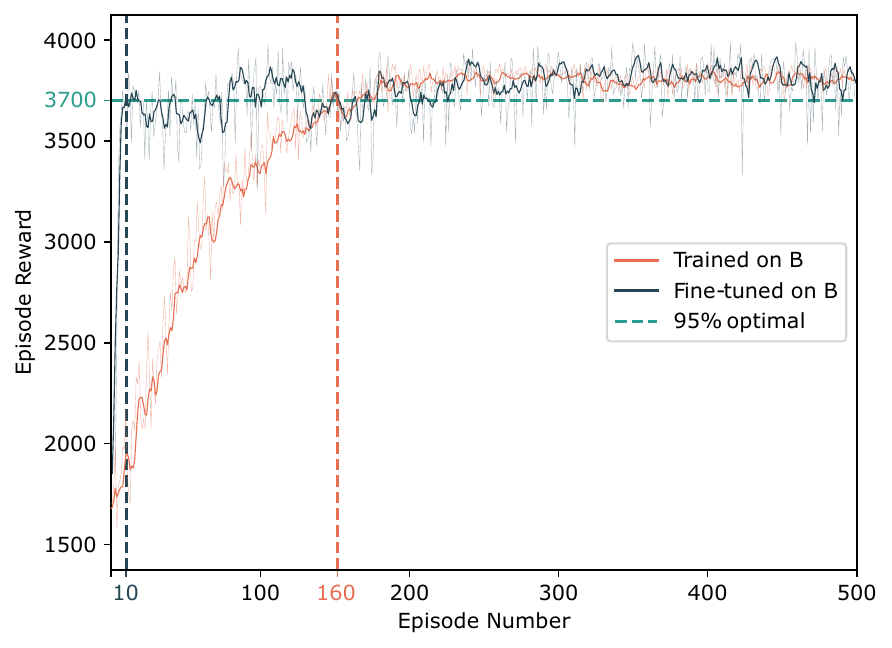}
    \caption{Comparison of the performance of training the \gls{rl} agent from scratch on environment B to the performance of an agent trained on A and fine-tuned on B. The green dashed line represents 95  $\%$ of the maximum reward achieved by the agent during the entire training process.}
    \label{fig:transfer_learning_effect}
\end{figure}

As shown in Fig.~\ref{fig:transfer_learning_effect}, two cases are compared: training a model on environment B from scratch (orange) and training a model on environment A followed by fine-tuning in environment B (dark blue). To achieve 95 $\%$ of the optimal performance (indicated by the green dashed horizontal line), the model must be trained for 160 episodes (shown by the orange vertical line) in environment B. However, achieving the same performance requires only 10 episodes (shown by the dark blue vertical line) of fine-tuning a pre-trained model on environment B, thereby reducing computational complexity and training time by a factor of 16.

Moreover, training the \gls{dqn} model in the simulation environment requires 26 M \gls{macop} operations per episode. Consequently, reaching 160 episodes demands 4251 M \gls{macop} operations. In contrast, with the transfer learning scheme, only 265 M \gls{macop} operations are needed, significantly reducing the computational load. On a machine with an Intel(R) Core(TM) i5-1145G7 2.60 GHz CPU, this translates to reducing the training time from 20 min to 1.2 min. In this work, only 200 training points per episode and a simple \gls{dqn} with 3 fully connected layers are used. In more realistic scenarios, both the number of training samples and the complexity of neural network architectures are greater, making the proposed scheme even more beneficial for significantly reducing training requirements.

Additionally, the comparison between the two training approaches highlights the stability and efficiency of the transfer learning method. The fine-tuned model not only reaches the 95 $\%$ optimal performance faster but also maintains a higher average episode reward throughout the training process, as evidenced by the smoother and more consistent reward curve in dark blue. This demonstrates that transfer learning can effectively leverage previously acquired knowledge to accelerate learning and improve the robustness of beam selection strategies in new environments.

\section{Conclusion}
This paper demonstrates the effectiveness of transfer learning for beam selection in 5G and beyond networks, achieving significant reductions in training time and computational complexity. By modeling the environment as a point cloud and leveraging the Chamfer distance to identify structurally similar scenarios, we enable efficient reuse of pre-trained \gls{rl} models across diverse propagation conditions. Simulation results confirm that our approach reduces training time and computational load by a factor of 16, while maintaining high performance. Beyond performance improvements, this work makes a concrete step toward sustainable \gls{ai} for wireless communications. By minimizing redundant training and accelerating time-to-deployment, the proposed method reduces energy consumption and lowers the environmental footprint of \gls{ai}-enabled \gls{ran} functions. This aligns with ongoing efforts to design greener, resource-aware communication systems.

Future work will explore the integration of our approach with Channel Knowledge Maps to enrich the point cloud representation with detailed channel characteristics. This fusion could further improve beam selection accuracy and efficiency, and contribute to the broader vision of scalable, adaptive, and energy-efficient \gls{ai} in next-generation wireless networks.

\bibliographystyle{IEEEtran}
\bibliography{references}

\end{document}